\documentclass{article}

\PassOptionsToPackage{breaklinks=true}{hyperref}

\PassOptionsToPackage{table}{xcolor}

\usepackage[final,main]{neurips_2026}

\usepackage{conference}
\usepackage{xcolor}
\usepackage{array}
\usepackage{needspace}
\definecolor{iprpbg}{RGB}{248,252,255}
\newcommand{\qcol}{\cellcolor{iprpbg}}

\title{Don't Waste the Population: Post-Anneal Refinement for Combinatorial Optimization}

\author{%
  Yuma Ichikawa\\
  Fujitsu Limited\\
  RIKEN Center for AIP
}

\hypersetup{pdftitle={Don't Waste the Population: Post-Anneal Refinement for Combinatorial Optimization},
  pdfauthor={Yuma Ichikawa},hidelinks}

\begin{document}
\maketitle

\begin{abstract}
    Parallel annealers generate diverse candidate solutions; yet, conventional post-processing typically refines only the best, leaving much of the population's potential untapped. We introduce Iterative Path-Relinking Polish (I-PRP), a deterministic, training-free method that exploits this diversity to discover better solutions without modifying the underlying annealing process. I-PRP selects promising candidates, explores cost-guided paths between locally optimized solutions, and refines intermediate states to uncover better local minima. Newly discovered improvements are iteratively incorporated into the population, enabling further exploration. We theoretically guarantee that I-PRP never degrades solution quality relative to conventional single-candidate polishing on the same population and converges in finitely many iterations. Experiments across challenging combinatorial optimization benchmarks, including MaxCut, MaxClique, and graph coloring, demonstrate substantial improvements in several instances. Paired evaluations confirm non-degradation, while matched-runtime comparisons show that exploiting existing population diversity can outperform additional annealing. These results establish I-PRP as an effective and reliable refinement strategy for parallel annealing.
\end{abstract}

\section{Introduction}\label{sec:intro}

Combinatorial optimization underlies scheduling, routing, and graph analysis, but many of its basic formulations are computationally hard~\citep{papadimitriou1998combinatorial,korte2011combinatorial,karp2010reducibility}. Practical solvers must balance solution quality with the computational cost of each new instance. Learning-based methods can amortize heuristic design across related instances~\citep{bengio2021machine,cappart2023combinatorial,li2018combinatorial,khalil2017learning,karalias2020erdos}, while training-free methods optimize each instance directly through sampling, relaxation, or local search~\citep{kirkpatrick1983optimization,lourenco2003iterated}. We focus on improving the discrete solution returned by an existing training-free pipeline.

Recent solvers expose more of their search to parallel hardware. Locally informed samplers use energy differences to guide discrete proposals~\citep{grathwohl2021oops,zhang2022langevin,sun2023discrete,sun2023revisiting,NEURIPS2023_f9ad87c1}. Continuous-relaxation annealers optimize soft variables before driving them toward discrete configurations~\citep{ichikawa2024controlling}. Parallel Quasi-Quantum Annealing (PQQA) combines this relaxation with a population of interacting replicas~\citep{ichikawa2025pqqa}. The resulting candidates can occupy different regions of the objective landscape.

The post-processing pipeline considered here retains only the best rounded PQQA replica and applies a greedy one-flip polish. This selection uses the population to choose a starting point but does not exploit relationships among the discarded candidates. Two rounded replicas can descend to different local optima even when their objective values are similar. Their coordinate disagreements then specify a route between those optima, along which further local searches may discover a better solution.

Path relinking explores such routes by gradually introducing the attributes of one reference solution into another~\citep{glover2000fundamentals,ribeiro2012path}. It has been combined with elite archives in greedy randomized adaptive search procedures (GRASP), tabu search, and quadratic optimization~\citep{festa2002randomized,marti2009advanced,wang2012path}. Here, the reference solutions are already available in the rounded annealing population. The question is how to use them as a deterministic refinement once the upstream search has finished.

We introduce \textbf{Iterative Path-Relinking Polish (I-PRP)}. A pass polishes the top rounded elites, constructs cost-guided paths from the best polished elite to the others, and polishes each intermediate state. These paths may move uphill; the returned incumbent changes only when a polished candidate improves its objective. A successful pass inserts that candidate into the elite tuple and repeats the search. We prove non-degradation relative to the baseline on the same rounded population and the finite termination of the exact iteration. At termination, none of the polished states on the final population's paths improves the incumbent. This certificate describes the routes actually explored and does not require the population to cover all local moves.

The experiments evaluate this refinement on the maximum independent set, maximum clique, maximum cut, and graph coloring. The reported gains are most pronounced on RBtest MaxClique and many large G-set MaxCut instances; categorical relinking also resolves the residual coloring conflicts on \texttt{anna} and \texttt{jean}. Paired controls on saved populations separate endpoint selection, intermediate re-polishing, and iteration. Comparisons with longer and restarted PQQA runs measure the value of this refinement under matched time budgets. Together, these results motivate retaining several basin representatives after annealing, even when only one solution is ultimately returned.

\section{Background}\label{sec:background}

For a positive integer $m$, write $[m]=\{1,\ldots,m\}$. Binary configurations lie in $\{0,1\}^N$, where $N$ is a positive integer, and $H(\B{x},\B{y})=|\{i\in[N]:x_i\ne y_i\}|$ denotes Hamming distance. All objective values are finite real numbers, and smaller values are better unless an experimental score is explicitly identified as a maximization objective.

\paragraph{Discrete objectives and penalties.}
For an instance $C$ with an original cost $f_0$ and $M$ constraints, a penalty formulation is \citep{lucas2014ising,kochenberger2014unconstrained}
\begin{equation}\label{eq:co-penalised}
 \min_{\B{x}\in\{0,1\}^N} l(\B{x};C,\B{\lambda}),\qquad
 l(\B{x};C,\B{\lambda})=f_0(\B{x};C)+\sum_{i=1}^{M}\lambda_i v_i(\B{x};C),
\end{equation}
where $\lambda_i>0$, $v_i\ge0$, and all penalties vanish exactly on feasible configurations. When interpreting global penalized optima as constrained optima, we assume that the feasible set is nonempty and every infeasible configuration has penalized cost strictly larger than the best feasible cost. The refinement guarantees below do not need this exact-penalty assumption: they hold for the fixed scalar objective $f(\B{x})=l(\B{x};C,\B{\lambda})$ itself. In particular, descent of a general penalty objective alone is not a feasibility guarantee.

A quadratic unconstrained binary optimization (QUBO) objective has the form $f(\B{x})=\B{x}^{\top}Q\B{x}+\B{q}^{\top}\B{x}$, with $Q=Q^{\top}\in\mab{R}^{N\times N}$ and $\B{q}\in\mab{R}^N$. Under $s_i=2x_i-1$ for $i\in[N]$, it can be written, up to an additive constant, as the Ising objective $f(\B{s})=-\B{s}^{\top}J\B{s}/2-\B{h}^{\top}\B{s}$, with $J=J^{\top}$ and $J_{ii}=0$. These representations share the same discrete search structure.

\paragraph{Graph objectives.}
The binary benchmarks differ in their discrete scores and penalty terms. For a simple graph $G=([N],E)$, an independent set selects vertices with no selected edge. A standard QUBO encoding is
\begin{equation}\label{eq:mis-objective}
 f_{\mathrm{MIS}}(\B{x})=-\sum_{i=1}^N x_i
 +\lambda\sum_{\{i,j\}\in E}x_ix_j,\qquad \lambda>1.
\end{equation}
The first term rewards cardinality and the second penalizes selected edges. With this strict penalty condition, every one-flip local optimum is feasible: if a selected vertex has $d\ge1$ selected neighbors, removing it changes the objective by $1-\lambda d<0$. On a feasible state, minimizing this objective is equivalent to maximizing the independent-set size. Maximum clique has the same encoding with the edge set replaced by the graph's nonedges. This gives the same feasibility argument for clique solutions. The strict condition matters; at $\lambda=1$, a violating move can be neutral and the strict polish need not remove it.

For a weighted graph, a cut assigns the two sides by $x_i\in\{0,1\}$. If edge $\{i,j\}$ has real weight $w_{ij}$, its cut value and minimization objective are
\begin{equation}\label{eq:cut-objective}
 c(\B{x})=\sum_{\{i,j\}\in E}w_{ij}(x_i+x_j-2x_ix_j),
 \qquad f_{\mathrm{cut}}(\B{x})=-c(\B{x}).
\end{equation}
Thus, a lower optimization objective corresponds to a larger reported cut. These encodings explain the relationship between local descent and the benchmark scores. The general I-PRP analysis depends only on the resulting fixed objective, so it also applies to other penalty choices; the feasibility conclusion above specifically requires the stated encoding and $\lambda>1$.

\paragraph{Parallel quasi-quantum annealing.}\label{subsec:bg-pqqa}
PQQA maintains $S\ge1$ relaxed replicas $\B{p}^{(s)}\in[0,1]^N$ and applies gradient-based sampling~\citep{ichikawa2025pqqa}. Its per-replica relaxed objective is
\begin{equation}\label{eq:pqqa-objective-main}
 \hat r(\B{p};C,\B{\lambda},\gamma)
 =\hat l(\B{p};C,\B{\lambda})+\gamma s(\B{p}),
 \qquad s(\B{p})=\sum_{i=1}^N\bigl[1-(2p_i-1)^4\bigr],
\end{equation}
where $\hat l$ is differentiable and agrees with $l$ at the binary vertices. The concave regularizer $s$ is largest at the half-integral vector and vanishes at the binary vertices. Annealing $\gamma$ from negative to positive first favors interior values and then penalizes them~\citep{ichikawa2024controlling}. A cross-replica term encourages population diversity.

The final replicas are rounded by $\widetilde{x}^{(s)}_i=1$ if $p^{(s)}_i\ge1/2$ and $0$ otherwise. Write the rounded population as $\widetilde{\mac{P}}=(\widetilde{\B{x}}^{(1)},\ldots,\widetilde{\B{x}}^{(S)})$. We order states by increasing objective, breaking ties lexicographically with $0<1$. The baseline post-processing in this paper polishes only the first state in this order. I-PRP uses the same rounded population and modifies this final refinement step.

\paragraph{One-flip polish and basins.}
Let $\sigma_i\B{x}$ denote $\B{x}$ with coordinate $i$ flipped, and define $\Delta_i(\B{x})=f(\sigma_i\B{x})-f(\B{x})$. For the symmetric QUBO,
\begin{equation}\label{eq:delta-qubo}
 \Delta_i^{\mathrm{QUBO}}(\B{x})
 =(1-2x_i)\bigl[q_i+Q_{ii}+2(Q\B{x})_i-2Q_{ii}x_i\bigr].
\end{equation}
The corresponding Ising difference is $\Delta_i^{\mathrm{spin}}(\B{s})=2s_i[(J\B{s})_i+h_i]$. Both can be computed from a cached matrix-vector product; Appendix~\ref{subsec:impl-oneflip} derives the differences and cache updates.

The deterministic map $\mathsf{Polish}:\{0,1\}^N\to\{0,1\}^N$ performs repeatedly.
\begin{equation}\label{eq:polish-update}
 \B{x}\leftarrow\sigma_{i^\ast}\B{x},\qquad
 i^\ast=\min\arg\min_{j\in[N]}\Delta_j(\B{x}),
 \qquad\text{while }\min_{j\in[N]}\Delta_j(\B{x})<0.
\end{equation}
The smallest-index rule resolves ties. Each accepted move strictly decreases $f$, so no state is revisited, and the finite search terminates. Its output is one-flip locally optimal, meaning that every one-flip difference is nonnegative. Consequently, $\mathsf{Polish}$ is idempotent. For each locally optimal state $\B{x}^{\circ}$, define its basin as
\begin{equation}
 \mac{B}(\B{x}^{\circ})
 =\{\B{x}\in\{0,1\}^N:\mathsf{Polish}(\B{x})=\B{x}^{\circ}\}.
\end{equation}
Determinism makes these basins a partition of the cube, including landscapes with tied objective values. They describe the outcome of this particular polish rather than an assumed geometric neighborhood. The baseline selects one basin; the remaining replicas can provide representatives of others.

\section{Method}\label{sec:method}

\begin{figure}[tb]
  \centering
  \includegraphics[width=0.98\linewidth]{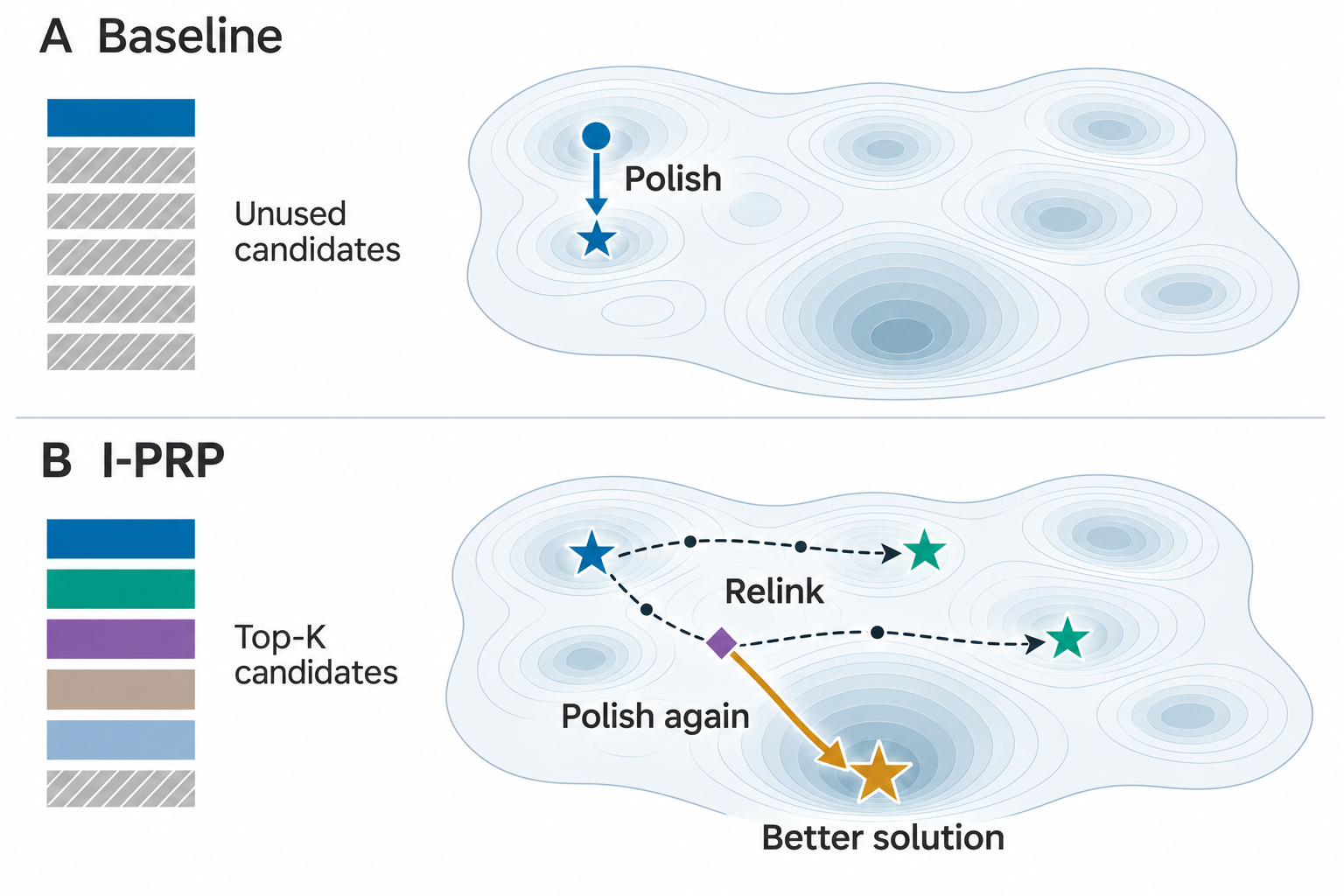}
  \caption{Post-anneal refinement. (A) The baseline polishes only the best rounded replica. (B) I-PRP polishes the top-$K$ elites and follows separate paths from the best polished elite to the others. Re-polishing an intermediate state can reach a better basin. The paths retain their endpoints throughout each pass; the landscape is schematic.}
  \label{fig:iprp-concept}
\end{figure}

I-PRP uses coordinate disagreements between polished elites to construct search paths (Figure~\ref{fig:iprp-concept}). The path may increase the objective temporarily, while a separate incumbent retains the best polished solution found. All formal results use complete paths, exact objective comparisons, and the deterministic polish defined above.

For two states $\B{x}_a,\B{x}_b\in\{0,1\}^N$, path relinking starts at $\B{c}^{(0)}=\B{x}_a$. At step $t$, let $D_t=\{i:c_i^{(t)}\ne x_{b,i}\}$. If $D_t\ne\emptyset$, choose the lowest-cost flip toward the target:
\begin{equation}\label{eq:pr-step}
    i_t^\ast=\min\arg\min_{j\in D_t}\Delta_j(\B{c}^{(t)}),
    \qquad \B{c}^{(t+1)}=\sigma_{i_t^\ast}\B{c}^{(t)}.
\end{equation}
Each move removes one coordinate from the disagreement set, so the path reaches $\B{x}_b$ after exactly $H(\B{x}_a,\B{x}_b)$ steps. Unlike local descent, the move need not improve $f$. This permits the path to leave a local optimum before a fresh polish is applied~\citep{glover2000fundamentals,ribeiro2012path}.

\subsection{One pass: Path-Relinking Polish}\label{subsec:method-prp}

Sort the rounded population by objective and lexicographic order, and take its top-$K$ prefix
\begin{equation}\label{eq:topk-prefix}
 \mac{P}=(\B{x}^{(1)},\ldots,\B{x}^{(K)}),\qquad 1\le K\le S.
\end{equation}
Repeated states are allowed. The one-pass map $\Phi_K:(\{0,1\}^N)^K\to\{0,1\}^N$ processes this tuple as follows.

\paragraph{Stage 1: polish the elites.}
Compute $\mathsf{Polish}(\B{x}^{(k)})$ for all $k\in[K]$, and sort these outputs by objective and lexicographic order. Denote the sorted polished states by $\B{x}^{(1)\star},\ldots,\B{x}^{(K)\star}$. Their indices refer to this new order, not to the raw input ranks. Initialize the incumbent to $\B{x}^{\star}=\B{x}^{(1)\star}$ and write $b(\mac{P})=f(\B{x}^{(1)\star})$. These states serve as basin representatives. Raw objective order need not agree with polished objective order, so this step alone can improve on polishing only the best rounded replica.

\paragraph{Stage 2: relink the polished endpoints.}
For $k=2,\ldots,K$ in that order, apply~\eqref{eq:pr-step} from $\B{x}^{(1)\star}$ to $\B{x}^{(k)\star}$. Write $\B{c}^{(t,k)}$ for its state after $t$ moves, $D_t^{(k)}$ for its disagreement set, and $d_k=|D_0^{(k)}|$. The endpoints remain fixed throughout the pass, even if another candidate improves the incumbent. Duplicate endpoints have $d_k=0$ and require no path moves.

\paragraph{Stage 3: polish the path states.}
After each path move, polish a copy of the visited state:
\begin{equation}\label{eq:stage3-repolish}
 \B{z}^{(t,k)}=\mathsf{Polish}(\B{c}^{(t,k)}),\qquad 1\le t\le d_k.
\end{equation}
Replace $\B{x}^{\star}$ by $\B{z}^{(t,k)}$ only when $f(\B{z}^{(t,k)})<f(\B{x}^{\star})$. Polishing does not alter the path state or its target. Return $\Phi_K(\mac{P})=\B{x}^{\star}$ after all paths have been processed. Strict promotion retains the earlier incumbent on ties, so the pass is deterministic.

Equivalently, the returned value is the minimum over the best polished elite and all polished path states:
\begin{equation}\label{eq:incumbent-minimum}
    f(\Phi_K(\mac{P}))
    =\min\left(\{b(\mac{P})\}\cup
    \left\{f\bigl(\mathsf{Polish}(\B{c}^{(t,k)})\bigr):
    2\le k\le K,\ 1\le t\le d_k\right\}\right).
\end{equation}
This identity separates the gain from polishing more elites from the further gain obtained by relinking. Its incumbent invariant is proved in Appendix~\ref{appx:proof-monotone}.

Re-polishing lets the search change coordinates outside the endpoint disagreement set. The path therefore supplies starting points for descents into basins that need not contain either endpoint. Appendix~\ref{appx:worked-example} illustrates this mechanism with a three-variable QUBO.

\begin{lemma}[Non-degradation on a fixed population]\label{lem:monotone}
    For every $K\ge1$ and every input tuple $\mac{P}=(\B{x}^{(1)},\ldots,\B{x}^{(K)})$,
    \begin{equation}
     f(\Phi_K(\mac{P}))\le\min_{k\in[K]} f(\mathsf{Polish}(\B{x}^{(k)})).
    \end{equation}
    If the tuple contains the best rounded replica, this also bounds the baseline polish on the same population.
\end{lemma}

The best polished elite is always available to the incumbent. Relinking can therefore improve its objective or leave it unchanged. This is a paired guarantee for a fixed population and objective; it does not compare independent annealing runs. If $K=1$, or all selected elites polish to the same state, I-PRP returns exactly the baseline polished state.

\begin{proposition}[Improvement along a two-step path]\label{prop:two-flip}
    Suppose $K\ge2$ and the polished states $\B{x}^{(1)\star}$ and $\B{x}^{(k)\star}$, for some $k\in\{2,\ldots,K\}$, disagree on exactly two distinct coordinates $\{i,j\}$. Let $i_0^\ast=\min\arg\min_{m\in\{i,j\}}\Delta_m(\B{x}^{(1)\star})$. If
    \begin{equation}
     f(\mathsf{Polish}(\sigma_{i_0^\ast}\B{x}^{(1)\star}))<f(\B{x}^{(1)\star}),
    \end{equation}
    then $f(\Phi_K(\mac{P}))<f(\B{x}^{(1)\star})$.
\end{proposition}

This is the shortest path with a non-endpoint state. Both endpoints are already polished, and the target is no better than the source. The intermediate state can nevertheless initiate a descent to a better basin. The condition concerns the coordinate actually selected by the deterministic rule; it does not assert that every improving two-flip move is explored. Proposition~\ref{prop:multi-flip} gives the corresponding statement for longer path prefixes.

\subsection{Iterating strict improvements: I-PRP}\label{subsec:method-iprp}

A successful pass adds a new basin representative to the tuple. Let $\operatorname{Sort}$ order a tuple by its raw objective values and lexicographic order. For a sorted tuple $\mac{P}$, define
\begin{equation}\label{eq:pop-update}
    \Psi_K(\mac{P})=
    \begin{cases}
    \operatorname{Sort}\bigl(\B{x}^{(1)},\ldots,\B{x}^{(K-1)},\Phi_K(\mac{P})\bigr),
     & f(\Phi_K(\mac{P}))<b(\mac{P}),\\
    \mac{P}, & \text{otherwise}.
    \end{cases}
\end{equation}
Thus, a strict improvement replaces the worst \emph{raw} member rather than the worst polished member. Starting from the top-$K$ rounded tuple $\mac{P}_0$, iterate $\mac{P}_{t+1}=\Psi_K(\mac{P}_t)$ until it is unchanged, then return $\Phi_K(\mac{P}_t)$. Algorithm~\ref{alg:iprp} provides the complete procedure.

\begin{theorem}[Finite convergence of I-PRP]\label{thm:iprp-fixed-point}
    For the exact iteration, let $b_t=b(\mac{P}_t)$ and $g_t=f(\Phi_K(\mac{P}_t))$. Both sequences are non-increasing. Every update that changes the tuple satisfies
    \begin{equation}
     b_{t+1}=g_t<b_t,\qquad g_{t+1}\le g_t.
    \end{equation}
    There are at most $|f(\{0,1\}^N)|-1\le2^N-1$ such updates. The iteration terminates at a tuple $\mac{P}^{\ast}$, and
    \begin{equation}
     \Psi_K(\mac{P}^{\ast})=\mac{P}^{\ast}
     \quad\Longleftrightarrow\quad
     f(\Phi_K(\mac{P}^{\ast}))=b(\mac{P}^{\ast}).
    \end{equation}
\end{theorem}

The inserted solution is already polished and is strictly better than every polished member of the old tuple. It therefore becomes the next pass's best polished elite. Since a finite objective range admits only finitely many strict decreases, the loop must stop. The bound proves termination; it does not provide a polynomial-time complexity guarantee. Lemma~\ref{lem:monotone} and the theorem also imply non-degradation relative to the initial baseline throughout the iteration.

\begin{corollary}[Optimality over the final relinking paths]\label{cor:two-flip-optimality}
    At a fixed point $\mac{P}^{\ast}$, let $\B{x}^{(1)\star,\ast},\ldots,\B{x}^{(K)\star,\ast}$ be its sorted polished elites. For each $k\in\{2,\ldots,K\}$, let $d_k^{\ast}=H(\B{x}^{(1)\star,\ast},\B{x}^{(k)\star,\ast})$ and $\B{c}^{(t,k),\ast}$ be the corresponding deterministic path state. Then
    \begin{equation}\label{eq:cor-intermediate-polish}
     f(\mathsf{Polish}(\B{c}^{(t,k),\ast}))\ge b(\mac{P}^{\ast}),
     \qquad 1\le t\le d_k^{\ast}.
    \end{equation}
\end{corollary}

The final pass has exhausted the routes specified by its elite population: none of their polished states improve the incumbent. This population-covered certificate does not imply optimality over all two-flip or higher-order neighborhoods. Complete proofs appear in Appendix~\ref{appx:proofs}.

\begin{algorithm}[tb]
\caption{Iterative Path-Relinking Polish (I-PRP).}
\label{alg:iprp}
\begin{algorithmic}[1]
\Require rounded population $\widetilde{\mac{P}}\in(\{0,1\}^N)^S$, objective $f$, integer $1\le K\le S$
\Ensure a polished state with objective no greater than polishing the best rounded member
\State sort $\widetilde{\mac{P}}$ by $(f,\mathrm{lex})$; let $\mac{Q}$ be its first $K$ entries
\Loop \label{ln:iprp-start}
  \State $(\B{x}^{(1)},\ldots,\B{x}^{(K)})\gets\mac{Q}$
  \State compute $\mathsf{Polish}(\B{x}^{(k)})$ for all $k\in[K]$ \label{ln:phi-start}
  \State sort these states by $(f,\mathrm{lex})$, naming them $\B{x}^{(1)\star},\ldots,\B{x}^{(K)\star}$
  \State $b\gets f(\B{x}^{(1)\star})$; $(\B{x}^{\star},f^{\star})\gets(\B{x}^{(1)\star},b)$
  \For{$k=2,\ldots,K$}
    \State $\B{c}^{(0,k)}\gets\B{x}^{(1)\star}$; $t\gets0$
    \State $D\gets\{i\in[N]:c_i^{(0,k)}\ne x_i^{(k)\star}\}$
    \While{$D\ne\emptyset$}
      \State $i^\ast\gets\min\{i\in\arg\min_{j\in D}\Delta_j(\B{c}^{(t,k)})\}$
      \State $\B{c}^{(t+1,k)}\gets\sigma_{i^\ast}\B{c}^{(t,k)}$; $D\gets D\setminus\{i^\ast\}$
      \State $\B{z}\gets\mathsf{Polish}(\B{c}^{(t+1,k)})$ \Comment{leave the path state unchanged}
      \If{$f(\B{z})<f^{\star}$}
        \State $(\B{x}^{\star},f^{\star})\gets(\B{z},f(\B{z}))$
      \EndIf
      \State $t\gets t+1$
    \EndWhile
  \EndFor \label{ln:phi-end}
  \If{$f^{\star}=b$}
    \State \Return $(\B{x}^{\star},f^{\star})$
  \EndIf
  \State $\mac{Q}\gets\operatorname{Sort}(\B{x}^{(1)},\ldots,\B{x}^{(K-1)},\B{x}^{\star})$
\EndLoop \label{ln:iprp-end}
\end{algorithmic}
\end{algorithm}

\paragraph{Cost and parallelism.}
In a dense QUBO, initializing a cached product costs $O(N^2)$; scanning the flip differences, copying a state and cache, or updating them after one flip costs $O(N)$. Let $F_k^{\mathrm{elite}}$ and $F_{t,k}^{\mathrm{path}}$ count accepted flips in the respective initial and path-state polish calls. Including worst-case lexicographic sorting, one pass costs
\begin{equation}\label{eq:complexity}
 O\!\left(KN^2+KN\log K
 +N\sum_{k=1}^{K}F_k^{\mathrm{elite}}
 +N\sum_{k=2}^{K}\sum_{t=1}^{d_k}(1+F_{t,k}^{\mathrm{path}})\right).
\end{equation}
Here $d_k\le N$, and path-polish calls receive copies of the current cache. Sparse matrix updates can use incident nonzeros, although a full coordinate scan still costs $O(N)$ without an additional data structure. Initial elite polishes and different target paths are independent within a pass. They may be batched, with their outputs reduced in the specified order to preserve deterministic ties; steps along an individual path remain sequential.

\section{Experiments}\label{sec:experiments}

We compare the improved sampling algorithm for combinatorial optimization (iSCO)~\citep{sun2023revisiting,NEURIPS2023_f9ad87c1}, baseline PQQA~\citep{ichikawa2025pqqa}, and PQQA followed by I-PRP. The suites include maximum independent set (MIS), maximum clique, and maximum cut (MaxCut) on Erd\H{o}s--R\'enyi (ER), Barab\'asi--Albert (BA), and random regular graphs (RRG), as well as established graph benchmarks. In the tables, $n$ is the vertex count and $d$ is the degree of a regular graph. For maximization tasks, the tables report a quality ratio $\rho$ (also called ApR) relative to the stated reference; higher is better. References need not be certified optima, so these ratios can exceed one. Graph coloring instead reports the number of monochromatic edges, where lower is better.

\paragraph{Setup.}
The benchmark tables use the PQQA annealing protocol, with per-suite settings in Appendix~\ref{subsec:impl-repro}. The core refinement uses $K=32$ on G-set and $K=8$ elsewhere, complete paths, and strict improvement. These tables summarize independently run pipelines. Section~\ref{subsec:exp-ablation} complements them with same-population and matched-runtime controls, directly testing the setting of Lemma~\ref{lem:monotone}. Synchronized component times are reported in Appendix~\ref{subsec:paired-timing}. Optional configurations and their same-archive ablation appear in Appendices~\ref{subsec:extended-benchmarks} and~\ref{subsec:paired-pr-controls}, separately from the core operator and its certificate.

\begin{table}[tb]
\centering
\caption{MIS and maximum-clique benchmark quality from independent runs. Columns report the mean reference ratio $\bar\rho$ and standard deviation $\sigma$; higher ratios are better. Bold marks the largest displayed mean in each row, including ties; shading identifies I-PRP. RRG ratios use the benchmark reference values~\citep{qqa4coBench}. $^\dagger$ Reported single-basin baseline value. Paired comparisons and synchronized times appear in Appendix~\ref{appx:paired-experiments}.}
\label{tab:qubo}
\setlength{\tabcolsep}{9pt}
\renewcommand{\arraystretch}{1.08}
\normalsize
\begin{tabular}{@{}lrrrrr>{\columncolor{iprpbg}[\the\tabcolsep][0pt]}r@{}}
\toprule
& \multicolumn{2}{c}{iSCO} & \multicolumn{2}{c}{PQQA} & \multicolumn{2}{c}{PQQA + I-PRP} \\
Suite & $\bar\rho$ & $\sigma$ & $\bar\rho$ & $\sigma$ & \qcol $\bar\rho$ & \qcol $\sigma$ \\
\midrule
\multicolumn{7}{@{}l}{\textit{Maximum independent set}} \\
SATLIB-UF ($n{\approx}1209$) & 0.987 & 0.003 & \textbf{0.994} & 0.002 & \cellcolor{iprpbg} \textbf{0.994}$^\dagger$ & \qcol 0.002 \\
ER $[700,800]$ & 0.926 & 0.014 & \textbf{0.998} & 0.007 & \cellcolor{iprpbg} 0.997 & \qcol 0.008 \\
ER $[9000,11000]$ & 0.865 & 0.007 & \textbf{0.995} & 0.003 & \cellcolor{iprpbg} \textbf{0.995} & \qcol 0.003 \\
\multicolumn{7}{@{}l}{\textit{Random regular MIS} (dense-$Q$)} \\
RRG $d{=}20,n{=}10^4$ & 0.679 & 0.001 & \textbf{0.757} & 0.001 & \cellcolor{iprpbg} 0.755 & \qcol 0.001 \\
RRG $d{=}100,n{=}10^4$ & 0.835 & 0.004 & 0.943 & 0.002 & \cellcolor{iprpbg} \textbf{0.949} & \qcol 0.003 \\
RRG $d{=}20,n{=}10^5$ & 0.591 & 0.001 & \textbf{0.753} & 0.000 & \cellcolor{iprpbg} \textbf{0.753} & \qcol 0.000 \\
RRG $d{=}100,n{=}10^5$ & 0.739 & 0.002 & \textbf{0.954} & 0.001 & \cellcolor{iprpbg} 0.953 & \qcol 0.001 \\
\multicolumn{7}{@{}l}{\textit{Maximum clique}} \\
Twitter & \textbf{1.000} & 0.000 & 0.999 & 0.009 & \cellcolor{iprpbg} \textbf{1.000} & \qcol 0.003 \\
RBtest & 0.880 & 0.050 & 0.879 & 0.053 & \cellcolor{iprpbg} \textbf{0.888} & \qcol 0.053 \\
\bottomrule
\end{tabular}
\end{table}

\subsection{QUBO suites: MIS, Max Clique, and MaxCut}\label{subsec:exp-qubo}

\paragraph{Maximum independent set.}
Table~\ref{tab:qubo} reports ratios on SATLIB-UF~\citep{hoos2000satlib}, ER, and regular graphs. The SATLIB-UF result matches baseline PQQA at $0.994$. Both PQQA configurations achieve ratios near the reference value on the two ER suites. On the high-degree regular graph with $d=100$ and $n=10^4$, I-PRP increases the reported ratio from $0.943$ to $0.949$. The regular-graph results use the dense QUBO representation and their stated benchmark references.

\paragraph{Maximum clique.}
On Twitter, I-PRP reaches a displayed mean ratio of $1.000$. On RBtest, the ratio increases from $0.879$ for baseline PQQA to $0.888$, compared with $0.880$ for iSCO. RBtest thus provides a setting in which retaining and relinking several polished candidates improves the aggregate solution quality.

\paragraph{Maximum cut.}
Table~\ref{tab:gset} reports the core operator on G-set. Its mean cut exceeds baseline PQQA on 15 of the 16 instances and exceeds iSCO on 11, with gains on G55--G57, G59--G62, and G64--G67. For example, the mean cut on G59 increases from $5865.6$ to $5966.8$, and on G64 from $8372.0$ to $8556.0$. These improvements show the value of refining a population after the annealing schedule has ended.

\begin{table}[tb]
    \centering
    \caption{Mean cut values on G-set~\citep{ye2003yyye}; higher is better. These are independent benchmark runs, rather than refinements of a shared population. Raw cuts permit direct comparison without a normalization reference. Bold marks the largest displayed mean in each row.}
    \label{tab:gset}
    \setlength{\tabcolsep}{23pt}
    \renewcommand{\arraystretch}{1.04}
    \normalsize
    \begin{tabular}{@{}lrr>{\columncolor{iprpbg}[\the\tabcolsep][0pt]}r@{}}
    \toprule
    Instance & iSCO & PQQA & PQQA + I-PRP \\
    \midrule
    G14 & \textbf{3054.6} & 2970.4 & \qcol 3044.6 \\
    G15 & \textbf{3042.8} & 2945.0 & \qcol 3035.6 \\
    G22 & \textbf{13312.2} & 13119.4 & \qcol 13279.4 \\
    G55 & 10165.0 & 10195.2 & \qcol \textbf{10209.8} \\
    G56 & 3873.6 & 3916.2 & \qcol \textbf{3951.8} \\
    G57 & 3398.4 & 3396.0 & \qcol \textbf{3428.0} \\
    G58 & \textbf{19104.2} & 18800.8 & \qcol 19064.8 \\
    G59 & 5875.6 & 5865.6 & \qcol \textbf{5966.8} \\
    G60 & 13952.4 & 14041.2 & \qcol \textbf{14049.0} \\
    G61 & 5555.4 & 5660.0 & \qcol \textbf{5684.6} \\
    G62 & 4719.2 & 4725.6 & \qcol \textbf{4771.2} \\
    G63 & \textbf{26749.4} & 26343.6 & \qcol 26706.2 \\
    G64 & 8388.6 & 8372.0 & \qcol \textbf{8556.0} \\
    G65 & 5356.0 & 5391.3 & \qcol \textbf{5393.3} \\
    G66 & 6106.7 & 6140.7 & \qcol \textbf{6151.3} \\
    G67 & 6660.0 & \textbf{6724.7} & \qcol 6721.3 \\
    \bottomrule
    \end{tabular}
\end{table}

\Needspace{12\baselineskip}
\subsection{Graph coloring with categorical I-PRP}\label{subsec:exp-coloring}

For a simple undirected graph $G=([N],E)$ and an integer $L\ge2$, an $L$-color configuration belongs to $[L]^N$ and the objective is
\begin{equation}\label{eq:coloring-objective}
 f(\B{x})=\sum_{\{u,v\}\in E}\mathbf{1}\{x_u=x_v\},
\end{equation}
where $\mathbf{1}\{A\}$ is $1$ when $A$ holds and $0$ otherwise. A local move changes a vertex $i$ to a color $a\ne x_i$. The categorical polish repeatedly selects the move with the smallest objective difference, breaks ties lexicographically by $(i,a)$, and accepts a move only when its difference is negative, stopping when none remains. A relinking step instead changes one disagreeing vertex directly to its target color, choosing the smallest objective difference and then the smallest vertex index. Each step removes one disagreement. The finite-descent and idempotence arguments, and hence Lemma~\ref{lem:monotone} and Theorem~\ref{thm:iprp-fixed-point}, apply on the finite state space $[L]^N$, with $L^N$ replacing $2^N$ in the state-count bound.

Table~\ref{tab:coloring} compares Tabucol, a tabu-search method, with PI-SAGE, the GraphSAGE variant of the physics-inspired graph neural network solver. I-PRP retains the zero-conflict solutions of PQQA and achieves proper colorings on \texttt{anna} and \texttt{jean}, reducing their reported conflicts from $3$ and $1$ to zero. The other reported conflict counts remain unchanged. This extension uses the same elite selection and strict promotion rules as the binary operator.

\begin{table}[tb]
    \centering
    \caption{Graph coloring conflicts with $L$ colors; lower is better. Tabucol~\citep{hertz1987using} and PI-SAGE reference scores are from~\citet{schuetz2022graphcoloring}. Mycielski rows use benchmark-specific identifiers and show only PQQA-based results. $L$ is the tested color budget; dashes denote omitted reference comparisons.}
    \label{tab:coloring}
    \setlength{\tabcolsep}{7pt}
    \renewcommand{\arraystretch}{1.0}
    \normalsize
    \begin{tabular}{@{}lcccc>{\columncolor{iprpbg}[\the\tabcolsep][0pt]}c@{}}
    \toprule
    Instance & $L$ & Tabucol & PI-SAGE & PQQA & \qcol \textbf{I-PRP} \\
    \midrule
    myciel3 & 3 & -- & -- & 0 & \qcol \textbf{0} \\
    myciel4 & 4 & -- & -- & 0 & \qcol \textbf{0} \\
    myciel5 & 6 & -- & -- & 0 & \qcol \textbf{0} \\
    myciel6 & 7 & -- & -- & 0 & \qcol \textbf{0} \\
    myciel7 & 8 & -- & -- & 0 & \qcol \textbf{0} \\
    queen5\_5 & 5 & 0 & 0 & 0 & \qcol \textbf{0} \\
    queen6\_6 & 7 & 0 & 0 & 0 & \qcol \textbf{0} \\
    queen7\_7 & 7 & 0 & 0 & 0 & \qcol \textbf{0} \\
    queen8\_8 & 9 & 0 & 1 & 0 & \qcol \textbf{0} \\
    queen9\_9 & 10 & 0 & 1 & 0 & \qcol \textbf{0} \\
    queen10\_10 & 11 & -- & -- & 1 & \qcol \textbf{1} \\
    queen11\_11 & 11 & 20 & 17 & 11 & \qcol \textbf{11} \\
    queen12\_12 & 12 & -- & -- & 13 & \qcol \textbf{13} \\
    queen13\_13 & 13 & 35 & 26 & 14 & \qcol \textbf{14} \\
    anna & 11 & -- & -- & 3 & \qcol \textbf{0} \\
    jean & 10 & -- & -- & 1 & \qcol \textbf{0} \\
    queen8\_12 & 12 & -- & -- & 0 & \qcol \textbf{0} \\
    \bottomrule
    \end{tabular}
\end{table}

\subsection{Paired attribution and compute controls}\label{subsec:exp-ablation}

The paired study fixes the rounded PQQA population for each instance and seed, saves its tensor, and verifies its SHA-256 hash across refinement variants. Table~\ref{tab:paired-main} compares core I-PRP with standard one-replica polish on 90 population--seed records. It yields 29 strict improvements and 61 ties, directly matching the pointwise setting of Lemma~\ref{lem:monotone}. The confidence intervals use $10{,}000$ percentile cluster-bootstrap resamples of instances, with all seeds of an instance retained together. Appendix~\ref{subsec:paired-protocol} gives the complete component comparison and a separate paper-scale ER-MaxCut control.

\begin{table}[tb]
    \centering
    \caption{Paired gains of core I-PRP over standard polish on identical rounded populations. Gains are in cut units on G-set and vertices on RBtest and RRG-MIS. ``Wins'' counts strict gains; all other records are ties. These targeted populations differ from the larger benchmark suites above.}
    \label{tab:paired-main}
    \small
    \setlength{\tabcolsep}{6pt}
    \renewcommand{\arraystretch}{1.08}
    \begin{tabular}{@{}lrrlr@{}}
    \toprule
    Suite & Pairs & Mean gain & 95\% confidence interval & Wins \\
    \midrule
    G-set G1--G5 & 15 & $+194.600$ & $[176.067,215.067]$ & 15/15 \\
    RBtest, first 20 instances & 60 & $+0.167$ & $[0.033,0.317]$ & 10/60 \\
    RRG-MIS ($d=100$, $n=1{,}000$) & 15 & $+0.267$ & $[0.067,0.467]$ & 4/15 \\
    \bottomrule
    \end{tabular}
\end{table}

\paragraph{Where the gain comes from.}
On the 15 G-set pairs, polishing the elite endpoints improves the mean cut by $55.133$ over standard polish. Relinking without intermediate re-polishing gives $72.533$; one complete re-polishing pass gives $141.667$; iteration gives $194.600$. These comparisons distinguish the two candidate sources in Eq.~\eqref{eq:incumbent-minimum}: a better polished endpoint and a better descent initiated along a path. Reinserting an improvement then changes the paths available to the next pass. On a separate six-population ER study, the core method obtains $84.4\%$ of the gain of the full optional extension stream. Same-archive forward, reverse, and bidirectional relinking controls are reported in Appendix~\ref{subsec:paired-pr-controls}.

\paragraph{Quality under matched time budgets.}
We use synchronized measurements on an NVIDIA GH200 and set the target budget to shared PQQA annealing followed by core I-PRP. On G1--G5, I-PRP exceeds longer PQQA by $213.533$ cuts and restarted PQQA by $115.600$ cuts on average, winning all 15 pairs in each comparison. The competitors use $0.970$ and $0.966$ of the target time. The advantage over longer PQQA is also present at each reported quarter-budget checkpoint. On the targeted RRG-MIS suite, I-PRP exceeds runtime-matched iSCO by $7.000$ vertices on average across 15 wins, at an iSCO-to-target time ratio of $0.985$. Appendix~\ref{subsec:paired-timing} reports the timing decomposition, uncertainty, and the RBtest controls.

\paragraph{Population size and returned solutions.}
The six-population ER sweep gives a mean gain of $20.667$ cuts at $K=8$, compared with $25.000$ at $K=64$, using $1.946$ rather than $25.491$ seconds of mean post-processing time. Thus a small archive captures most of the gain observed at $K=64$ in this study. Appendix~\ref{subsec:paired-sensitivity} separates this fixed-population comparison from reruns that change the upstream replica count or diversity coefficient. Independent checks on 15 ER populations find no improving two-flip or three-flip move in any returned I-PRP state; the corresponding pass counts are 2/15 for standard polish and 5/15 for one pass. These are empirical properties of the tested outputs, while Corollary~\ref{cor:two-flip-optimality} remains a certificate over the final population's paths.

\Needspace{12\baselineskip}
\section{Related work}\label{sec:related-work}

\paragraph{Path relinking and elite archives.}
Path relinking uses reference solutions to guide trajectories through discrete search spaces~\citep{glover2000fundamentals}. Its combinations with GRASP, tabu search, and quadratic optimization are well established~\citep{festa2002randomized,resende2004grasp,ribeiro2012path,marti2009advanced,wang2012path}. Other effective local and population search approaches include breakout search, tabu search, and multilevel refinement~\citep{benlic2013breakout,kochenberger2013solving,wang2012multilevel}. Our contribution concerns where the reference set comes from and how the refinement stops: the final rounded PQQA population supplies the archive, and strict improvements drive a deterministic outer iteration. The method does not require an archive to be maintained during annealing.

\paragraph{Training-free optimization.}
Annealing, Markov chain Monte Carlo, and locally informed samplers explore low-cost states without supervised training~\citep{kirkpatrick1983optimization,neal1996sampling,hukushima1996exchange,earl2005parallel,angelini2019monte,grathwohl2021oops,zhang2022langevin}. iSCO is a discrete sampling baseline in the PQQA benchmark protocol~\citep{sun2023revisiting,NEURIPS2023_f9ad87c1}. Continuous relaxations and PQQA instead expose differentiable objectives to parallel computation~\citep{ichikawa2024controlling,ichikawa2025pqqa}. I-PRP uses only their final discrete candidates, a scalar objective, and a deterministic local polish; the upstream sampling or relaxation dynamics do not enter its guarantees.

\paragraph{Learning-based solvers.}
Graph neural networks and neural relaxations learn or optimize heuristics for combinatorial tasks~\citep{li2018combinatorial,khalil2017learning,karalias2020erdos,schuetz2022combinatorial,sun2022annealed,xu2020tilingnn,cappart2023combinatorial,bengio2021machine}. Studies of physics-inspired graph neural networks also emphasize comparisons with established search methods~\citep{angelini2023modern,boettcher2023inability,schuetz2023reply}. I-PRP requires no model training. Its interface could also refine a population generated by a learned solver, provided the objective and deterministic polish are supplied.

\section{Conclusion}\label{sec:conclusion}

The final population of a parallel annealer can support further searches after its best candidate has been selected. I-PRP uses polished elites to define relinking paths and new local descents, retaining a new solution only under strict improvement. For the same rounded input, it does not worsen the baseline objective, terminates in a finite number of updates, and exhausts the polished candidates on the final population's paths. The QUBO and categorical results illustrate how this refinement can improve solutions while leaving the annealing loop unchanged.

\paragraph{Scope and limitations.}
I-PRP depends on the supplied objective, penalty weights, and population. Relinking provides no gain beyond endpoint polishing if the elites share one basin or their paths do not exhibit an improving descent. Refinement adds problem-dependent computation; the elite-size sweep and synchronized timings quantify this quality–time trade-off. Its certificate concerns only the final elite-induced paths and guarantees neither strict improvement nor global optimality. The finite-state upper bounds are exponential. Optional candidate generators and truncated-path variants are evaluated separately from the complete-path certificate.

\paragraph{Broader impact.}
Improved combinatorial search may support scheduling and resource allocation. Its downstream effects depend on the chosen objectives and constraints; optimizing an incomplete model can reinforce undesirable allocation decisions. The method itself does not assess fairness or the suitability of a deployment objective.

\clearpage
\begin{ack}
\textbf{Funding.} The author received no external funding or other third-party
support for this work.

\textbf{Competing interests.} The author declares no competing interests.
\end{ack}

\bibliographystyle{plainnat}
\bibliography{ref}

\newpage
\appendix
\section{Algorithmic Details}\label{appx:algorithm}

Algorithm~\ref{alg:iprp} gives the exact iteration in the main paper. The following example tracks all of its decisions, including the path-state copy, the raw-rank replacement, and the final stopping test.

\subsection{A complete three-variable example}\label{appx:worked-example}

Consider the QUBO
\begin{equation}
 f(\B{x})=2x_1+3x_2+2x_3-4x_1x_2-5x_1x_3+4x_2x_3,
 \qquad \B{x}\in\{0,1\}^3,
\end{equation}
and the initial elite tuple $\mac{P}_0=(000,110)$, writing binary
triples without commas. Both elites are strict one-flip local minima:
their objective values are $0$ and $1$, and their flip differences are
$(2,3,2)$ and $(2,1,1)$, respectively. Stage~1, therefore, leaves them
unchanged. The target $110$ differs from $000$ in coordinates $1$ and
$2$. Since their initial flip costs are $2$ and $3$, the first relinking
path is
\begin{equation}
 000 (f=0)\ \longrightarrow\ 100 (f=2)
 \ \longrightarrow\ 110 (f=1).
\end{equation}
At $100$, the flip differences are $(-2,-1,-3)$, so polishing flips
coordinate $3$ and reaches $101$ with objective $-1$. Its flip
differences are $(3,3,3)$, so it is also a strict local minimum.
Thus $\Phi_2(\mac{P}_0)=101$: polishing leaves the relinking path by
changing a coordinate on which the endpoints agree.

The update replaces the worst raw elite $110$, giving
$\mac{P}_1=(101,000)$. The second pass relinks these two states.
Both differing coordinates have flip cost $3$, so the index rule selects coordinate $1$:
\begin{equation}
 101 (f=-1)\ \longrightarrow\ 001 (f=2)
 \ \longrightarrow\ 000 (f=0).
\end{equation}
Polishing $001$ returns $101$, because its flip differences are
$(-3,7,-2)$; polishing $000$ leaves it unchanged. Neither candidate
improves the incumbent value $-1$. Hence $\mac{P}_1$ is a fixed point
and I-PRP returns $101$ after one strict population update.

\section{Proofs}\label{appx:proofs}

Throughout, $f:\{0,1\}^N\to\mab{R}$ is fixed, comparisons are exact, and every relinking path is complete. The smallest-index rule makes $\mathsf{Polish}$ deterministic. Each accepted flip strictly decreases $f$, ensuring that the finite state space guaranties termination. The stopping rule implies one-flip local optimality and, consequently, idempotence. For a tuple $\mac{P}$, write $b(\mac{P})=f(\B{x}^{(1)\star})$ and $d_k=H(\B{x}^{(1)\star},\B{x}^{(k)\star})$. The state $\B{c}^{(t,k)}$ is the $t$-th state on the path to elite $k$; all endpoints remain fixed during a pass.

\subsection{Proof of Lemma~\ref{lem:monotone}}\label{appx:proof-monotone}

\begin{proof}
    The incumbent starts at the smallest polished-elite value $b(\mac{P})$ and is replaced precisely when a candidate has a strictly smaller value. Induction over the finite sequence of evaluated candidates proves Eq.~\eqref{eq:incumbent-minimum}.
    The first set is nonempty even when there are no path states. Since sorting does not change the polished values,
    \begin{equation}
    f(\Phi_K(\mac{P}))\le b(\mac{P})
    =\min_{1\le k\le K}f(\mathsf{Polish}(\B{x}^{(k)})).
    \end{equation}
    For the top-$K$ prefix of the same sorted rounded population, $\B{x}^{(1)}$ is the upstream choice, yielding the paired non-degradation claim. Every possible returned state is a polish output and hence is itself polished.
\end{proof}

\subsection{Proof of Proposition~\ref{prop:two-flip}}\label{appx:proof-two-flip}

\begin{proof}
    Fix the elite $k$ in the hypothesis. Its disagreement set is $\{i,j\}$, so the deterministic path rule gives
    $\B{c}^{(1,k)}=\sigma_{i_0^\ast}\B{x}^{(1)\star}$.
    Stage~3 evaluates this state, and Eq.~\eqref{eq:incumbent-minimum} implies
    \begin{equation}
    f(\Phi_K(\mac{P}))
    \le f\bigl(\mathsf{Polish}(\sigma_{i_0^\ast}\B{x}^{(1)\star})\bigr)
    < f(\B{x}^{(1)\star}).
    \end{equation}
    The endpoint is already polished and has value at least $f(\B{x}^{(1)\star})$ by sorting. Thus the asserted improvement is supplied by the polished intermediate, or by an even better candidate evaluated elsewhere in the pass.
\end{proof}

\subsection{Proof of Theorem~\ref{thm:iprp-fixed-point}}\label{appx:proof-theorem}

\begin{proof}
Write $\mac{P}_t=(\B{x}_t^{(1)},\ldots,\B{x}_t^{(K)})$, and let $b_t=b(\mac{P}_t)$ and $g_t=f(\Phi_K(\mac{P}_t))$. Lemma~\ref{lem:monotone} gives $g_t\le b_t$. The returned state $\B{y}_t=\Phi_K(\mac{P}_t)$ is polished, so
\begin{equation}\label{eq:phi-output-polished}
\mathsf{Polish}(\B{y}_t)=\B{y}_t,
\qquad f(\mathsf{Polish}(\B{y}_t))=g_t.
\end{equation}
If $g_t=b_t$, the tuple is unchanged, and determinism gives $b_{t+1}=b_t$ and $g_{t+1}=g_t$.

Suppose $g_t<b_t$. Every retained raw member has a polished value of at least $b_t$, whereas the inserted member has a polished value of $g_t$. Therefore
\begin{equation}\label{eq:btp1-full}
b_{t+1}
=\min\left(\{g_t\}\cup
\{f(\mathsf{Polish}(\B{x}_t^{(k)})):1\le k<K\}\right)
=g_t<b_t.
\end{equation}
Moreover, $\B{y}_t$ cannot already be a member of $\mac{P}_t$: idempotence would then give a polished-elite value of $g_t<b_t$, contradicting the definition of $b_t$. Thus, insertion changes the tuple even after sorting. Applying Lemma~\ref{lem:monotone} to the next tuple yields
\begin{equation}
g_{t+1}\le b_{t+1}=g_t.
\end{equation}
Both sequences are consequently non-increasing, with a strict decrease of $b_t$ at every update.

The set $f(\{0,1\}^N)$ has at most $2^N$ elements. Hence, there are at most $|f(\{0,1\}^N)|-1$ updates. Each pass also terminates, since it contains finitely many path moves and terminating polish calls. The iteration therefore reaches a fixed tuple $\mac{P}^\ast$. For any sorted tuple, $f(\Phi_K(\mac{P}))\le b(\mac{P})$; equality selects the unchanged branch, while strict inequality changes the tuple as just proved. This establishes the claimed fixed-point characterization in both directions.
\end{proof}

\subsection{Proof of Corollary~\ref{cor:two-flip-optimality}}\label{appx:proof-corollary}

\begin{proof}
At a fixed tuple, Theorem~\ref{thm:iprp-fixed-point} gives $f(\Phi_K(\mac{P}^\ast))=b(\mac{P}^\ast)$. By Eq.~\eqref{eq:incumbent-minimum}, every evaluated candidate must then have a value of at least $b(\mac{P}^\ast)$, which is exactly Eq.~\eqref{eq:cor-intermediate-polish}. 
\end{proof}

\subsection{Longer relinking paths}\label{appx:proof-multi-flip}

\begin{proposition}[Improvement from a path prefix]\label{prop:multi-flip}
Let $K\ge2$ and $k\in\{2,\ldots,K\}$. Suppose the polished elites $\B{x}^{(1)\star}$ and $\B{x}^{(k)\star}$ differ on a set $D\subseteq[N]$ of size $m\ge2$. Let $\pi:[m]\to D$ be the bijection listing the coordinates selected in order by their deterministic Stage~2 path. For $1\le r<m$, define
\begin{equation}\label{eq:y-r}
\B{y}^{(r)}=\sigma_{\pi(r)}\cdots\sigma_{\pi(1)}\B{x}^{(1)\star}.
\end{equation}
If, for some such $r$,
\begin{equation}\label{eq:multi-flip-hyp}
f\bigl(\mathsf{Polish}(\B{y}^{(r)})\bigr)<f(\B{x}^{(1)\star}),
\end{equation}
then $f(\Phi_K(\mac{P}))<f(\B{x}^{(1)\star})$.
\end{proposition}

\begin{proof}
Each path move flips one coordinate that disagrees with the target, making that coordinate agree without changing any other coordinate. Thus the path selects every element of $D$ exactly once. Induction on $t$ gives
\begin{equation}\label{eq:path-induction-detailed}
\B{c}^{(t,k)}=\sigma_{\pi(t)}\cdots\sigma_{\pi(1)}\B{x}^{(1)\star},
\qquad D_t^{(k)}=D\setminus\{\pi(1),\ldots,\pi(t)\},
\end{equation}
for $1\le t\le m$. In particular, $\B{c}^{(r,k)}=\B{y}^{(r)}$. Equation~\eqref{eq:incumbent-minimum} and the hypothesis imply
\begin{equation}
f(\Phi_K(\mac{P}))
\le f\bigl(\mathsf{Polish}(\B{y}^{(r)})\bigr)
< f(\B{x}^{(1)\star}),
\end{equation}
as required.
\end{proof}

At $t=m$, the path reaches $\B{x}^{(k)\star}$, whose polished value is at least that of $\B{x}^{(1)\star}$. Only a proper prefix can therefore satisfy Eq.~\eqref{eq:multi-flip-hyp}. The prefix itself may be uphill; the improvement is obtained by polishing it. Proposition~\ref{prop:two-flip} is the case $m=2$ and $r=1$.

\section{Additional Implementation Details}\label{appx:impl-details}

This appendix gives the local objective updates, experimental settings, and optional candidate generators.

\subsection{One-flip differences and cache updates}\label{subsec:impl-oneflip}

For a symmetric QUBO objective $f(\B{x})=\B{x}^{\top}Q\B{x}+\B{q}^{\top}\B{x}$, let $\B{e}_i$ be the $i$-th standard basis vector, set $d_i=1-2x_i\in\{-1,1\}$, and write $\B{x}'=\sigma_i\B{x}=\B{x}+d_i\B{e}_i$. Expanding the objective difference gives
\begin{align}
\Delta_i^{\mathrm{QUBO}}(\B{x})
&= (\B{x}+d_i\B{e}_i)^{\top}Q(\B{x}+d_i\B{e}_i)+\B{q}^{\top}(\B{x}+d_i\B{e}_i)
   -\B{x}^{\top}Q\B{x}-\B{q}^{\top}\B{x} \notag\\
&= d_i\B{e}_i^{\top}Q\B{x}+d_i\B{x}^{\top}Q\B{e}_i+d_i^2Q_{ii}+d_iq_i \notag\\
&= 2d_i(Q\B{x})_i+Q_{ii}+d_iq_i, \label{eq:delta-qubo-expanded}
\end{align}
where symmetry of $Q$ and $d_i^2=1$ are used in the last line. The formula in the main text follows by writing
\begin{align}
(1-2x_i)&\bigl[q_i+Q_{ii}+2(Q\B{x})_i-2Q_{ii}x_i\bigr] \notag\\
&= d_iq_i+2d_i(Q\B{x})_i+d_iQ_{ii}(1-2x_i)
 = d_iq_i+2d_i(Q\B{x})_i+Q_{ii}. \label{eq:delta-qubo-general}
\end{align}
When the linear term is folded into the diagonal by $Q_{ii}\leftarrow Q_{ii}+q_i$ and $\B{q}=0$, this reduces to the diagonal-folded formula used in many QUBO implementations. A non-symmetric matrix can be replaced by $(Q+Q^{\top})/2$ without changing $\B{x}^{\top}Q\B{x}$, so the symmetry assumption is not restrictive for QUBO objectives.

For Ising objectives $f(\B{s})=-1/2\B{s}^{\top}J\B{s}-\B{h}^{\top}\B{s}$ with symmetric $J$ and $J_{ii}=0$, flipping spin $s_i$ changes only terms incident to $i$. Since the interaction contribution involving $i$ is $-s_i\sum_{j\ne i}J_{ij}s_j$, the flip changes it to $+s_i\sum_{j\ne i}J_{ij}s_j$. The field term changes from $-h_is_i$ to $+h_is_i$. Thus
\begin{equation}\label{eq:delta-ising-appx}
\Delta_i^{\mathrm{spin}}(\B{s})
=2s_i\sum_{j\ne i}J_{ij}s_j+2h_is_i
=2s_i\bigl[(J\B{s})_i+h_i\bigr].
\end{equation}
The cached products are updated in $O(N)$ per accepted flip in dense form. If $\B{x}'=\B{x}+d_i\B{e}_i$, then $Q\B{x}'=Q\B{x}+d_iQ\B{e}_i$. If spin $i$ is flipped, then $J\B{s}'=J\B{s}-2s_iJ\B{e}_i$, using the spin value before the flip. Thus a polish call needs one initial cached product, then scans all one-flip differences and updates the product after each accepted flip~\citep{merz2002greedy}.

\subsection{Experimental configuration and reproduction essentials}\label{subsec:impl-repro}

The original benchmark runs follow the PQQA hyperparameters of~\citet{ichikawa2025pqqa}. Their default settings use AdamW with learning rate $0.1$, entropy exponent $\alpha=4$, Langevin temperature $T=10^{-3}$, and communication-term coefficient $\alpha_{\mathrm{div}}=0.2$. For these benchmarks, I-PRP uses $K=32$ on G-set and $K=8$ elsewhere, with complete paths and strict improvement. The paired protocols and sensitivity variations are specified separately in Appendix~\ref{appx:paired-experiments}.

\begin{table}[h]
\centering
\caption{PQQA hyperparameters per problem class.}\label{tab:hp}
\small
\begin{tabular}{lrrrr}
\toprule
Suite & $S$ (replicas) & Steps & $\gamma_{\min}$ & $\gamma_{\max}$ \\
\midrule
MaxCut Optsicom, ER, BA, G14, G15  & 1000 & 30000 & $-5$  & 0.1 \\
MaxCut G-set G22 ($n{=}2{,}000$)   &  400 & 30000 & $-20$ & 0.1 \\
MaxCut G-set G55 ($n{=}5{,}000$)   &  200 & 30000 & $-20$ & 0.1 \\
MIS SATLIB-UF, MIS ER ($n \in [700,800]$)  & 1000 & 30000 & $-2$  & 0.1 \\
MIS RRG ($n = 10^4, d \in \{20, 100\}$)    &  200 & 30000 & $-2$  & 0.1 \\
Max Clique Twitter, RBtest                 & 1000 & 30000 & $-2$  & 0.1 \\
\bottomrule
\end{tabular}
\end{table}

The benchmark tables report aggregate quality scores; Table~\ref{tab:qubo} also gives the reported standard deviations. QUBO and coloring suites use the seed sets and upstream benchmark splits of~\citet{ichikawa2025pqqa}. For batched suites with a common maximum dimension, instances are grouped along the instance axis; G-set and the $n=10^4$ regular-graph MIS suite use a single-instance pipeline because their dimensions vary substantially. The additional paired studies in Appendix~\ref{appx:paired-experiments} use saved populations and explicitly reported sample counts. Their synchronized component timings separate post-processing cost from the shared anneal.

\subsection{Optional dense-MaxCut extensions}\label{subsec:impl-extensions}

The formal results in Section~\ref{sec:method} apply to the bare deterministic $\Phi_K/\Psi_K$ operator. The MaxCut configurations in Table~\ref{tab:extended-benchmarks} enable an optional guarded extension stack. Each generated candidate is promoted only if it strictly improves the incumbent, but the extended candidate stream is not part of Theorem~\ref{thm:iprp-fixed-point} or Corollary~\ref{cor:two-flip-optimality}.

\paragraph{Multi-flip path relinking.}
The optional path step changes one or two disagreeing coordinates at once. For distinct coordinates $i,j$ in a symmetric QUBO, with $d_i=1-2x_i$ and $d_j=1-2x_j$, its two-flip difference is
\begin{equation}
 f(\sigma_i\sigma_j\B{x})-f(\B{x})
 =\Delta_i(\B{x})+\Delta_j(\B{x})+2d_id_jQ_{ij}.
\end{equation}
This follows by expanding the quadratic form at $\B{x}+d_i\B{e}_i+d_j\B{e}_j$. Candidate promotion still requires strict improvement of the incumbent.

\paragraph{Consensus-guided subproblem solving.}
Coordinates on which all elites agree are fixed at their shared values. Substituting those values into the QUBO yields a quadratic objective on the remaining coordinates. For a sufficiently small remaining set, all binary assignments are enumerated and the best candidate is promoted only if it improves the incumbent.

\paragraph{Guarded restart and tabu candidates.}
Additional candidates are obtained from perturbed population members and tabu-search moves involving up to two variables. They are used in the optional MaxCut configurations and the extension stream in Appendix~\ref{subsec:paired-pr-controls}, with the same strict incumbent guard.

\subsection{Benchmark results with optional extensions}\label{subsec:extended-benchmarks}

Table~\ref{tab:extended-benchmarks} records the MaxCut configurations that use the optional candidate generators of Appendix~\ref{subsec:impl-extensions}. They are separate from the core comparisons in the main text. These entries come from independent end-to-end runs; their elapsed times do not measure the incremental overhead of adding I-PRP to a shared anneal. Table~\ref{tab:paired-timing} supplies that component comparison, and Table~\ref{tab:paired-pr-controls} isolates the extension stream on a common archive.

\begin{table}[tb]
\centering
\caption{MaxCut benchmark configurations with optional extensions. Columns report mean reference ratio $\bar\rho$, standard deviation $\sigma$, and recorded mean end-to-end time $t$. Bold marks the largest displayed ratio. The extended PQQA configuration includes candidates beyond the theorem-covered operator. DISCS reference scores are not certified optima~\citep{NEURIPS2023_f9ad87c1}.}
\label{tab:extended-benchmarks}
\small
\setlength{\tabcolsep}{4pt}
\renewcommand{\arraystretch}{1.06}
\begin{tabular}{@{}lrrrrrrrr>{\columncolor{iprpbg}[\the\tabcolsep][0pt]}r@{}}
\toprule
& \multicolumn{3}{c}{iSCO} & \multicolumn{3}{c}{PQQA} & \multicolumn{3}{c}{PQQA + extensions} \\
Suite & $\bar\rho$ & $\sigma$ & $t$ & $\bar\rho$ & $\sigma$ & $t$ & \qcol $\bar\rho$ & \qcol $\sigma$ & \qcol $t$ \\
\midrule
Optsicom & \textbf{1.000} & 0.000 & 1.1m & 0.999 & 0.004 & 13.2s & \cellcolor{iprpbg} \textbf{1.000} & \cellcolor{iprpbg} 0.003 & \qcol 15.0s \\
ER $n{=}256$--$300$ & \textbf{1.018} & 0.017 & 1.1m & 0.926 & 0.020 & 1.2m & \cellcolor{iprpbg} 1.017 & \cellcolor{iprpbg} 0.017 & \qcol 29.9s \\
BA $n{=}256$--$300$ & \textbf{1.000} & 0.000 & 1.3m & 0.988 & 0.004 & 1.2m & \cellcolor{iprpbg} 0.999 & \cellcolor{iprpbg} 0.002 & \qcol 50.5s \\
ER $n{=}512$--$600$ & \textbf{1.002} & 0.001 & 3.9s & 0.883 & 0.003 & 20.5s & \cellcolor{iprpbg} 1.000 & \cellcolor{iprpbg} 0.002 & \qcol 11.5s \\
BA $n{=}512$--$600$ & \textbf{1.010} & 0.005 & 4.0s & 0.982 & 0.006 & 20.5s & \cellcolor{iprpbg} 1.007 & \cellcolor{iprpbg} 0.006 & \qcol 46.4s \\
ER $n{=}1024$--$1100$ & \textbf{1.002} & 0.001 & 6.4s & 0.903 & 0.002 & 22.5s & \cellcolor{iprpbg} 1.000 & \cellcolor{iprpbg} 0.001 & \qcol 23.8s \\
BA $n{=}1024$--$1100$ & \textbf{1.017} & 0.003 & 6.6s & 0.982 & 0.004 & 22.4s & \cellcolor{iprpbg} 1.014 & \cellcolor{iprpbg} 0.004 & \qcol 1.6m \\
\bottomrule
\end{tabular}
\end{table}

\clearpage
\begingroup
\setlength{\intextsep}{8pt plus 2pt minus 2pt}
\section{Additional Experiments}\label{appx:paired-experiments}

\subsection{Shared populations and uncertainty}\label{subsec:paired-protocol}

The paired experiments hold the rounded PQQA population fixed for each instance and seed. The effective population tensor is saved and checked by its SHA-256 hash before each post-processing comparison. We distinguish the standard one-replica polish ($M_0$), the best of the polished top-$K$ endpoints ($M_1$), cost-guided relinking without intermediate re-polishing ($M_2$), one complete relinking-and-polish pass ($M_3$), the iterative core I-PRP operator ($M_4$), and the optional guarded extension stream ($M_5$). Thus $M_3$ evaluates $\Phi_K$ once, while $M_4$ repeats the strict-improvement updates until termination. The fixed-point certificate concerns $M_4$.

Gains are paired differences in the native maximization score: cut value for MaxCut, clique size for RBtest, and independent-set size for RRG-MIS. A positive difference favors I-PRP. The reported 95\% confidence intervals use the percentile cluster bootstrap with $10{,}000$ resamples of instances, retaining all seeds within each sampled instance. Win/tie/loss counts compare paired final scores.

\begin{table}[H]
\centering
\caption{Post-processing gains on the same rounded populations, relative to $M_0$. G-set uses G1--G5; RBtest uses its first 20 instances; RRG-MIS uses $d=100$ and $n=1{,}000$. The final interval is for the mean $M_4$ gain, and ``wins'' counts its strict gains over $M_0$.}
\label{tab:paired-components}
\small
\setlength{\tabcolsep}{3.5pt}
\begin{tabular}{@{}lrrrrrrl@{}}
\toprule
Suite & Pairs & $M_1$ & $M_2$ & $M_3$ & $M_4$ & Wins & 95\% CI \\
\midrule
G-set   & 15 & $+55.133$ & $+72.533$ & $+141.667$ & $+194.600$ & 15/15 & $[176.067,215.067]$ \\
RBtest  & 60 & $+0.000$  & $+0.083$  & $+0.167$   & $+0.167$   & 10/60 & $[0.033,0.317]$ \\
RRG-MIS & 15 & $+0.000$  & $+0.000$  & $+0.267$   & $+0.267$   & 4/15  & $[0.067,0.467]$ \\
\bottomrule
\end{tabular}
\end{table}

Table~\ref{tab:paired-components} contains 90 matched population--seed records. I-PRP improves 29 and preserves the baseline score in the other 61. On G-set, the progression from endpoint selection to relinking, intermediate re-polishing, and iteration isolates the contribution of each component. A separate six-pair ER-MaxCut comparison uses $S=1{,}000$, $30{,}000$ PQQA steps, and $K=8$: $M_4$ improves on $M_0$ by $16.667$ cuts on average, with six wins and a 95\% interval of $[4.667,29.833]$.

\subsection{Synchronized timing and matched-runtime controls}\label{subsec:paired-timing}

Component timings use an NVIDIA GH200 and the same PQQA anneal for all post-processing variants. GPU components are bracketed by synchronization calls, and warm-up is excluded. Table~\ref{tab:paired-timing} separates the shared anneal and rounding from the standard polish and the additional core refinement. The entries are component medians, rather than differences between independently timed pipelines.

\begin{table}[H]
\centering
\caption{Synchronized component times in seconds. Each entry is a median on the paired suites of Table~\ref{tab:paired-components}.}
\label{tab:paired-timing}
\small
\setlength{\tabcolsep}{8pt}
\begin{tabular}{@{}lrrr@{}}
\toprule
Suite & Anneal and round & Standard polish & Additional I-PRP \\
\midrule
G-set   & 1.480 & 0.04280 & 62.600 \\
RBtest  & 1.446 & 0.00118 & 0.145 \\
RRG-MIS & 1.493 & 0.00178 & 2.051 \\
\bottomrule
\end{tabular}
\end{table}

For the matched-runtime comparisons, the target is the time for the shared PQQA anneal followed by core I-PRP. Longer PQQA runs, independent PQQA restarts, and iSCO are calibrated separately on the same hardware. Table~\ref{tab:matched-runtime} reports the actual competitor time divided by this target. The ratios make the extent of budget matching explicit; the RBtest restart control uses $0.893$ of the target time.

\begin{table}[H]
\centering
\caption{I-PRP relative to runtime-matched controls. Gains use the native units of Table~\ref{tab:paired-components}. W/T/L counts I-PRP wins, ties, and losses. $r_t$ is competitor time divided by target time; dashes indicate intervals not reported for the iSCO comparisons.}
\label{tab:matched-runtime}
\small
\setlength{\tabcolsep}{3.5pt}
\begin{tabular}{@{}llrlrr@{}}
\toprule
Suite & Control & Mean gain & 95\% CI & W/T/L & $r_t$ \\
\midrule
G-set & Longer PQQA & $+213.533$ & $[187.667,237.533]$ & 15/0/0 & 0.970 \\
G-set & PQQA restarts & $+115.600$ & $[97.600,130.067]$ & 15/0/0 & 0.966 \\
RBtest & Longer PQQA & $+0.083$ & $[-0.033,0.217]$ & 11/43/6 & 0.973 \\
RBtest & PQQA restarts & $+0.167$ & $[0.033,0.317]$ & 10/50/0 & 0.893 \\
RRG-MIS & iSCO & $+7.000$ & -- & 15/0/0 & 0.985 \\
RBtest & iSCO & $+0.067$ & -- & 9/46/5 & 0.973 \\
\bottomrule
\end{tabular}
\end{table}

On G-set, I-PRP exceeds longer PQQA at all four measured budget checkpoints (Table~\ref{tab:paired-anytime}).

\begin{table}[H]
\centering
\caption{Best-so-far cut differences between I-PRP and longer PQQA on G1--G5 at fractions of the target budget. The 15 pairs are the same as in Table~\ref{tab:matched-runtime}.}
\label{tab:paired-anytime}
\small
\setlength{\tabcolsep}{7pt}
\begin{tabular}{@{}rrlr@{}}
\toprule
Budget & Mean gain & 95\% CI & W/T/L \\
\midrule
25\% & $+155.867$ & $[135.667,176.067]$ & 15/0/0 \\
50\% & $+190.400$ & $[169.267,209.337]$ & 15/0/0 \\
75\% & $+204.467$ & $[181.400,223.333]$ & 15/0/0 \\
100\% & $+213.533$ & $[187.667,237.533]$ & 15/0/0 \\
\bottomrule
\end{tabular}
\end{table}

\subsection{Relinking controls and optional extensions}\label{subsec:paired-pr-controls}

A six-population ER-MaxCut study uses edge probability $p=0.1$, $n\in\{100,200\}$, and three seeds per size. Every method in Table~\ref{tab:paired-pr-controls} receives the same saved archive. The directional controls perform cost-guided relinking with final polishing only; the bidirectional control considers both directions. Core I-PRP additionally re-polishes intermediate states and iterates after a strict gain.

\begin{table}[H]
\centering
\caption{Same-archive ER-MaxCut controls. Gains are relative to standard polish $M_0$; times are mean post-processing times. The final row includes optional candidate generators outside the core fixed-point certificate.}
\label{tab:paired-pr-controls}
\small
\setlength{\tabcolsep}{6pt}
\begin{tabular}{@{}lrr@{}}
\toprule
Post-processing method & Mean cut gain & Time (s) \\
\midrule
Endpoint polish only ($M_1$) & $+6.333$ & 0.028 \\
Forward relinking, final polish & $+15.167$ & 0.221 \\
Reverse relinking, final polish & $+15.000$ & 0.215 \\
Bidirectional relinking, final polish & $+16.500$ & 0.401 \\
Core iterative I-PRP ($M_4$) & $+20.667$ & 1.937 \\
Guarded extension stream ($M_5$) & $+24.500$ & 2.754 \\
\bottomrule
\end{tabular}
\end{table}

Core I-PRP attains $84.4\%$ of the full extension stream's gain using $70.3\%$ of its post-processing time in this experiment. The directional comparisons also show an additional gain from the complete I-PRP design when the input archive is held fixed.

\subsection{Elite size and upstream sensitivity}\label{subsec:paired-sensitivity}

Table~\ref{tab:paired-k-sweep} varies $K$ on the same six saved ER populations. Increasing $K$ exposes more potential endpoints, with a corresponding increase in refinement time. Here $K=8$ yields $82.7\%$ of the gain at $K=64$ while taking $7.6\%$ of its post-processing time.

\begin{table}[H]
\centering
\caption{Elite-size sensitivity on six fixed ER-MaxCut populations. Gains are relative to $M_0$; each value of $K$ is a separate core I-PRP run on the same saved inputs.}
\label{tab:paired-k-sweep}
\small
\setlength{\tabcolsep}{14pt}
\begin{tabular}{@{}rrrr@{}}
\toprule
$K$ & Mean cut gain & Strict gains & Mean time (s) \\
\midrule
1  & $0.000$ & 0/6 & 0.008 \\
2  & $+15.833$ & 6/6 & 0.190 \\
4  & $+18.667$ & 6/6 & 0.745 \\
8  & $+20.667$ & 6/6 & 1.946 \\
16 & $+24.333$ & 6/6 & 6.099 \\
32 & $+24.333$ & 6/6 & 15.425 \\
64 & $+25.000$ & 6/6 & 25.491 \\
\bottomrule
\end{tabular}
\end{table}

Table~\ref{tab:paired-other-sweeps} separates path caps from upstream changes. The cap $T_{\max}$ limits the relinking steps per path; the certificate in Corollary~\ref{cor:two-flip-optimality} covers complete paths. The upstream sweeps rerun PQQA at each population size $S$ or diversity coefficient. Every tested setting improves its paired baseline.

\begin{table}[H]
\centering
\caption{Post-processing and true-upstream sensitivity. Fixed-archive settings use six paired records; upstream settings use three. Mean gains are in cut units relative to the paired standard polish.}
\label{tab:paired-other-sweeps}
\small
\setlength{\tabcolsep}{8pt}
\begin{tabular}{@{}lrrr@{}}
\toprule
Parameter & Value & Mean gain & Strict gains \\
\midrule
\multicolumn{4}{@{}l}{\textit{Fixed saved populations}} \\
Path cap $T_{\max}/N$ & 0.25 & $+20.000$ & 6/6 \\
 & 0.50 & $+20.667$ & 6/6 \\
 & 1.00 & $+20.667$ & 6/6 \\
\midrule
\multicolumn{4}{@{}l}{\textit{PQQA rerun at each setting}} \\
Population size $S$ & 8  & $+14.000$ & 3/3 \\
 & 16 & $+14.000$ & 3/3 \\
 & 32 & $+7.000$ & 3/3 \\
 & 64 & $+9.667$ & 3/3 \\
Diversity coefficient $\alpha_{\mathrm{div}}$ & 0.00 & $+9.667$ & 3/3 \\
 & 0.05 & $+7.667$ & 3/3 \\
 & 0.10 & $+7.333$ & 3/3 \\
 & 0.20 & $+9.667$ & 3/3 \\
 & 0.40 & $+12.333$ & 3/3 \\
\bottomrule
\end{tabular}
\end{table}

\subsection{Fixed-point diagnostics}\label{subsec:paired-fixed-point}

In a small exact-enumeration study with 45 populations at $N\in\{12,14,16\}$, core I-PRP reaches a global optimum in all 45 cases and improves the standard polish in 22. A separate set of 15 ER-MaxCut populations at $N\in\{100,200,300\}$ is exhaustively checked for improvements through two-flip and three-flip moves. Table~\ref{tab:paired-fixed-point} reports the number of returned states passing each check. These are empirical neighborhood tests, separate from the population-covered guarantee.

\begin{table}[H]
\centering
\caption{Exhaustive neighborhood diagnostics on 15 ER-MaxCut populations. An entry counts returned states with no strictly improving move of the stated size.}
\label{tab:paired-fixed-point}
\small
\setlength{\tabcolsep}{12pt}
\begin{tabular}{@{}lrrr@{}}
\toprule
Check & Standard polish $M_0$ & One pass $M_3$ & Iterated $M_4$ \\
\midrule
Two-flip  & 2/15 & 5/15 & 15/15 \\
Three-flip & 2/15 & 5/15 & 15/15 \\
\bottomrule
\end{tabular}
\end{table}

The diagnostics also record distinct rounded states and polished basins, Hamming distances, path lengths, improving intermediates, and strict outer updates. The sufficient conditions in \mbox{Propositions}~\ref{prop:two-flip} and~\ref{prop:multi-flip} describe when polishing an intermediate state yields a better basin.

\paragraph{AI assistance.}
AI tools assisted with the experimental workflow, manuscript preparation, and schematic Figure~\ref{fig:iprp-concept}. They are not part of the optimization operator or its guarantees.
\endgroup


\end{document}